\documentclass[preprint,12pt]{elsarticle}
\usepackage{amssymb}
\usepackage{amsthm}
\usepackage{amsmath}
\usepackage{color}
\usepackage{multirow}
\usepackage{array}

\newcommand\abs[1]{\left|#1\right|}
\DeclareMathOperator*{\argmax}{arg\,max}
\DeclareMathOperator*{\MSE}{MSE}
\DeclareMathOperator*{\mean}{mean}

\usepackage{graphicx}
\usepackage{amssymb}
\usepackage{lineno}

\journal{XXX}

\begin{document}

\begin{frontmatter}

%% Title, authors and addresses

\title{Adaptive Experimental Design for Path-following Performance Assessment of Unmanned Vehicles}

%% use the tnoteref command within \title for footnotes;
%% use the tnotetext command for the associated footnote;
%% use the fnref command within \author or \address for footnotes;
%% use the fntext command for the associated footnote;
%% use the corref command within \author for corresponding author footnotes;
%% use the cortext command for the associated footnote;
%% use the ead command for the email address,
%% and the form \ead[url] for the home page:
%%
%% \title{Title\tnoteref{label1}}
%% \tnotetext[label1]{}
%% \author{Name\corref{cor1}\fnref{label2}}
%% \ead{email address}
%% \ead[url]{home page}
%% \fntext[label2]{}
%% \cortext[cor1]{}
%% \address{Address\fnref{label3}}
%% \fntext[label3]{}

%% use optional labels to link authors explicitly to addresses:
\author[label1,label2]{Eleonora~Saggini}
\author[label1]{Eva~Riccomagno}
\author[label2]{Massimo~Caccia}
\author[label3]{Henry~P.~Wynn}
\address[label1]{Department of Mathematics, University of Genova, Italy}
\address[label2]{Institute of Intelligent Systems for Automation, National Research Council, Italy}
\address[label3]{Department of Statistics, London School of Economics, UK}

%\author{Eleonora~Saggini,
%         Eva~Riccomagno,
%         Massimo~Caccia
%         and~Henry~P.~Wynn}

%\address{Department of Mathematics, University of Genova, Italy}

\begin{abstract}
%% Text of abstract
The definition of Good Experimental Methodologies (GEMs) in robotics is a topic of widespread interest due also to the increasing employment of robots in everyday civilian life. The present work contributes to the ongoing discussion on GEMs for Unmanned Surface Vehicles (USVs). It focuses on the definition of GEMs and provides specific guidelines for path-following experiments. 
Statistically designed experiments (DoE) offer a valid basis for developing an empirical model of the system being investigated.
A two-step adaptive experimental procedure for evaluating path-following performance and based on DoE, is tested on the simulator of the Charlie USV.
The paper argues the necessity of performing extensive simulations prior to the execution of field trials. 
\end{abstract}

\begin{keyword}
%Marine robotics \sep adaptive designs\sep motion control \sep performance metrics \sep underactuated robots ?? 
Robotics \sep Autonomous Vehicles \sep  Performance Monitoring \sep Statistical Design of Experiments %maximum of 6 keywords, using American spelling

%% keywords here, in the form: keyword \sep keyword

%% MSC codes here, in the form: \MSC code \sep code
%% or \MSC[2008] code \sep code (2000 is the default)

\end{keyword}

\end{frontmatter}

%%
%% Start line numbering here if you want
%%

%%%%%%%%%%%%%%%%%%%%%%%%%%%%%%%%%%%%%%%
%\linenumbers

%% main text
\section{Introduction}
\thispagestyle{empty}
\label{sec:intro}
In the robotic community there is an ongoing discussion on replicable and measurable experiments and on the definition of benchmarks for the performance evaluation of robots by means of experimental tests.
% There are a number of recent publication, e.g. the special issue~\cite{bonsignorio2015toward}, and 
One of the motivations of this discussion is the lack of laws and regulations to ensure the safe, reliable and effective use of robots in contexts where they can interact with humans~\cite{pobilIROS2006,Amigoni2007}.  
The consolidation of such technologies requires some additional efforts~\cite{bonsignorio2014fostering}, even though robotics has already been successfully introduced in daily life, e.g. in home automation and assistive applications that are taking place in the home and in industrial and service employments. 

In marine robotics, operating capabilities and technical solutions are available and partly in place. But various issues related to reliable methodology for executing operations, such as path-following, obstacle detection and avoidance, and the issue of safety and legal framework for the use of Unmanned Marine Vehicles (UMVs), prevent their use in civilian applications.  %Particularly in the marine robotics context, 
 But the total integration of UMVs in everyday life within civilian scenarios, i.e. in areas not restricted to maritime traffic, is desirable for many reasons. Some examples are the employment of UMVs as support to maintenance and operation of aquaculture plants, and within industrial contexts,  e.g. for underwater pipeline inspection and repetitive oceanographic measurements.
The lack of regulation has implied self regulation from all involved. For example the Legal Working Group of the Society's Autonomous Underwater Vehicles examined the legal status of AUVs in three volumes: the Report on the Law in $1999$, the Recommended Code of Practice in $2000$, and  The Law Governing AUV Operations -- Questions and Answers in $2001$
http://www.sut.org/ which are not universally accepted and have to be reviewed to account for the newest technologies.

Our work has implications in the reduction of the delay in technology transfer from research frameworks toward actual applicative scenarios. This delay is partly caused by the lack of standardized shared procedures for execution of experiments and comparison of  results~\cite{Bonsignorio2009}. 
Researchers in GEMs are actively working on creating a common ground and shared tools for easily replicating experiments already conducted by other researchers, for comparing results, and employing available data sets and common testing frameworks~\cite{Amigoni2009}. 
With the aim of obtaining reproducibility for the test campaigns and maximizing the relevance of their output, guidelines for {\em optimally} designing experiments are very much needed, especially in the field of marine robotics. This is heavily affected
by experimental constraints such as uncontrollability of external conditions, e.g., waves, sea currents, recreational and commercial traffic, by a restricted number of executable experiments due to cost and logistic issues, and by uncertainty in the robot inputs since hydrodynamic interactions, forces, and torques assigned to the system are inherently uncertain.
 
Considerations throughout the paper are tailored to surface marine robots, but the methodology presented in the paper equally applies to mobile robots on land, underwater marine vehicles executing path-following tasks at constant depth and any other scenarios involving the evaluation of the performance  while executing tasks along a one-dimensional curve. 
This paper is part of a research project on the development of Good Experimental Methodologies (GEMs) for testing and comparing Unmanned Marine Vehicles (UMVs). The project is carried out at the CNR-ISSIA and involves researchers with expertise in different areas including Control Engineering and Statistics. This article contributes to the project by discussing methods for assessing robots' performance and for planning experiments at sea and which can draw on practice in the statistical Design of Experiments (DoE). 

%The present work focuses on the definition of GEMs and specific guidelines for  path-following experiments. It argues the necessity of performing extensive simulations prior to the execution of trials at sea. 
The remainder of the paper is organized as follows:  Section~\ref{subsec:introBib} is a brief review of the literature on benchmarking in robotics for UMVs. In  Section~\ref{sec:experimentDefinition} a procedure for executing repeatable path-following experiments is presented. In  Section~\ref{sec:performanceIndices} the performance indices adopted by the research group to which the authors belong, are defined, while the standard mathematical models that approximate the performance functions are given in Section~\ref{sec:dace}. 
In Section~\ref{sec:doe} the theory is exemplified with two cases of 
path-following experiments and tested on the Charlie simulator~\cite{caccia2009charlie}. 
Results are reported in Section~\ref{sec:sim}. 
Charlie is a $2.40$~$m$-long and $1.7$~$m$-wide USV weighing $250$~$kg$. It was developed in the CNR-ISSIA laboratory and used for sampling operations in polar expeditions.

%---------------------------------------------------------------------
\section{Related work and our contribution}
\label{subsec:introBib}
The interest of the robotic community in defining and spreading GEMs has grown mainly in the last years. This is also due to the increasing variety of control algorithms which have been successfully tested to ensure the satisfactorily functionalities of the mechanics of the vehicles.
This research trend is testified by the establishment of interest
groups, e.g. Euron GEM Sig, by technical committees, e.g. IEEE Robotics and Automation TCPEBRAS16, 
by special issue in international journals~\cite{specialIssueRam2015,specialIssueAutonRobot2009},
workshops, e.g. ICRA 2011, 2012 and IEEE/RSJ IROS from 2006 to 2015,
and by an increasing number of publications by different authors~\cite{sprunk2014experimental, bonsignorio2015toward,  pobilIROS2006, Amigoni2007, bonsignorio2014fostering, Bonsignorio2009, Amigoni2009}.

A brief but comprehensive review of issues related to measuring and comparing research results in robotics is provided in~\cite{pobilIROS2006}, which includes also considerations on how to define  benchmarks in robotics and discusses the need of defining benchmarks for specific sub-domains of robotics rather than benchmarks valid for all domains.  
A detailed description of the main principles of the experimental methodology is given  in~\cite{Amigoni2009}.
In~\cite{Amigoni2007} the issues on how to perform, replicate and compare experiments are addressed for  robotic mapping, 
while in~\cite{Bonsignorio2009} general guidelines are proposed to improve methodologies and reporting. %in order to facilitate experiment replication,  performance evaluation and comparison for experiments in robotics. 
Topics on experimentation in mobile robot localization and mapping are discussed from an interdisciplinary viewpoint in~\cite{Amigoni2009}, touching on issues that stand at the crossroad of mobile robotics and philosophy of science. 
Field experiments for a Remotely Operated Vehicle (ROV) and an USV identification are reported in~\cite{Caccia2008JOE} and~\cite{Miskovic2011JFR}, respectively. Both articles cover the execution of experiments at sea, in presence of significant constraints in terms of controllability of the experimental conditions, the restricted number of executable experiments and the uncertainty in the inputs assigned to the system. 

Our group has addressed the definition of GEMs for path-following tasks from an engineering perspective in~\cite{Saggini2014performance,Saggini2015evaluation, Sorbara2015,saggini2014assessing}. 
During path-following the vehicle has to follow a predefined target path without any time constraints. This is a key task for marine robots, and in general mobile robots, because, from simple operations to more complex missions, many activities involve path-following executions. 
The performance indices and the methodology for analysing experiments presented in~\cite{Saggini2014performance} have been followed by other researchers.
%, thus testifying the increasing interest in this topic and encouraging the authors to continue working in this direction.  For instance, 
In~\cite{caharija2015comparison} the authors present and compare two different guidance algorithms on an eight-shape path. They consider cross track error measurements and servo command signals and restrict the analysis on the steady state phases. 
In~\cite{nadj2015navigation} an innovative overactuated USV capable of omnidirectional motion is presented: together with standard specifications about the mathematical model, the navigation, guidance and control (NGC) structure, the paper includes an analysis of the path-following performance of the vehicle on a square. 
However, due to a number of practical limitations, such measurements and analyses are limited to data gathered from a small number of tests. 

The choice of the methodology for investigating the capability of a vehicle in the execution of a task deserves attention and further investigation.
Relying on classical DoE methods and on simulation experiments would allow us to optimally select the tests to be executed at sea with great saving of time and resources, and increased quality of data and information on vehicle performance. 
Applications of DoE in control theory and robotics appear in various papers. A survey of methods from DoE relevant to control engineering is given in~\cite{pronzato2008optimal}
An application in mobile robotics for optimal allocation of sensors
can be found in~\cite{ucinski2000optimal},  while in~\cite{moreno2014optimal} it is described a method for the computation of the optimal trajectories which a surface vehicle should follow for tracking and positioning some  target vehicles. 
All those papers adopt model dependent strategies. Finally we cite~\cite{lizama1996designing} for an example of $G$-optimal experiments for dynamics identification. 
We exploit the classical theory of DoE and define a methodology for the choice of the paths for efficient gathering and analysis of experimental data during path following experimentations. 

%---------------------------------------------------------------------
\section{Path-following experiments for UMVs}
\label{sec:experimentDefinition}
In~\cite{sprunk2014experimental, bonsignorio2015toward,  pobilIROS2006, Amigoni2007, bonsignorio2014fostering, Bonsignorio2009, Amigoni2009} the need of GEMs in robotics is advocated. 
In marine robotics the practical possibility of guaranteeing the desirable reproducibility of results is limited by logistic and environmental constraints~\cite{Amigoni2009, Sorbara2015}. 
The experiments are greatly influenced by environmental constraints, such as waves, wind, sea currents,  and by other external disturbances, such as other vehicles, that unavoidably limit the degree of reproducibility of experiments at sea. 
Various research groups in Italy~\cite{bonsignorio2015toward,Sorbara2015}, Croatia~\cite{Miskovic2015RAM} and Spain~\cite{perez2015exploring} are facing this issue, encouraging the development of protocols and procedures for repeatable experiments, and for sharing the collected experimental data with the larger community of involved researchers,  engineers and users. 
The lack of standardized experimental procedures causes difficulty even in achieving tasks apparently simple, like driving a marine robot in a predefined position~\cite{Sorbara2015}. 
In marine robotics replicability refers to all controllable parameters of the experiments and not necessarily to their outcomes.

In this context, the authors have defined a protocol for executing replicable path-following experiments~\cite{Sorbara2015}. The interest in path-following is motivated by its large employment as a sub-task for other missions, e.g. obstacle avoidance, area coverage/sampling, and by the fact that it has typically a smoother convergence to the path with respect to similar tasks, such as reference-tracking.
The protocol in~\cite{Sorbara2015} is integrated in a software framework called \emph{DeepRuler}, a freely available tool, at the moment upon request to the authors. \emph{DeepRuler} allows the \emph{automatic} design, execution, monitoring and evaluation of path-following experiments. It can be installed on the UMV or run from remote. 

 \begin{figure*}[tb]
 \centering
 \includegraphics[width=\textwidth]{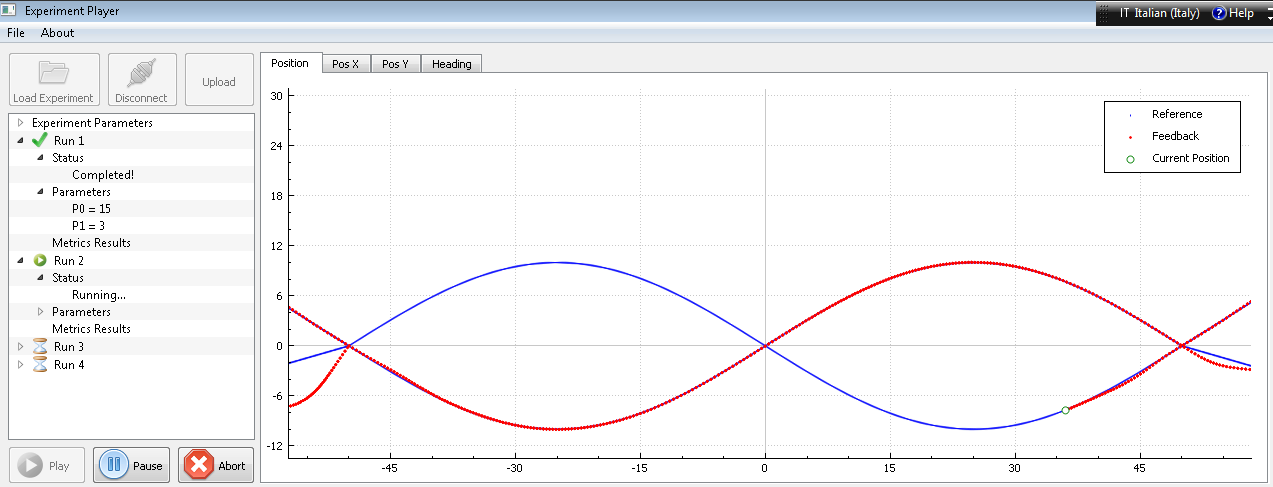}
 \caption{\label{fig:hci-player}The \emph{DeepRuler} human--computer interface player shows online simulation of the Charlie USV executing a sinusoidal path-following in backward direction. The red curve is the observed path and the blue curve the reference path.}
 \end{figure*}

Path-following experiments implemented by \emph{DeepRuler} depend on a minimal number of assumptions in order to make the protocol largely usable. In an experiment or batch there are $n$ runs and each run corresponds to a target (or ideal or reference) path that the vehicle should follow. Paths are executed sequentially in time and can be different within the same batch, although in practice this occurs rarely. 
At least ideally, we would like runs to be executed in such a way that they are independent. 
But this is very unlikely to happen during on field tests e.g. wind, currents, sea state will all be strongly correlated and will correlate the runs. 
We partially account for this in the modelling phase described in Section~\ref{sec:dace}, tie it in with the protocol for repeatability described in the next sentence, and test independence a-posteriori on the collected data. 
Precaution is required when comparing performances using tests executed in different environmental conditions. 
A run is divided into four sequential phases: approach, forward path, turn, backward path, so that each target path is executed under some environmental conditions and their opposite for reasonably short runs.

The experimenter is required to set some numerical values which will depend on the characteristic of manoeuvrability of the vehicle and on other constraints.
These include the edges $R_1$ and $R_2$ of a rectangular box where the vehicle can manoeuvre and the width $W$ of a central $W\times R_2$  rectangle box in which the forth and back paths are executed. 
Repeatability is achieved if the vehicle approaches this central box on a straight line in  the direction of the tangent to the target in its entry point into the central box. 
This applies to both back and forth paths. 
The vehicle turns outside the central box for example by performing a semi-circle of radius $r$ followed by a line of length $l$. 
This usually is not relevant because performance is measured only while the vehicle is in the central box. 
Eventually one would like \emph{DeepRuler} to choose automatically the values $R_1, R_2, W, l$ and $r$ according to the vehicle's characteristics, e.g. dimensions and turning radius, and to the morphology of the environment. The $n$ target paths in an experiment can be from a small dimensional parametric class of functions, such as sinusoids, circles, ellipses, or can be interpolating curves through some specified points. 
The choice of the type of paths to execute during a field trial is still being discussed and perfectioned. 
Again ideally we would like such choice to be made automatically by the system~\cite{lekkas2014guidance}. 
This is not implemented yet but does not affect the subject of the remainder of the paper. 
\emph{DeepRuler} is modular and new classes of target paths, either vehicle dependent or not, are included easily.  
The current version includes a sinusoidal path generation function in which each run is parametrized by the amplitude of the sine wave and by the number of half periods to be performed in a forth path. 
The straight line path, given as zero half periods, should be included in any experiment and performance indices could be defined relative to it. 
Figure~\ref{fig:hci-player} shows an experiment configured with ten runs. The active run is parametrized by an amplitude of ten meters and two half periods.

A reference path is assigned to the vehicle as a time series
 $\mathcal{R} = \{(x_{R,i},y_{R,i})$, $i = 1,\dots,n_R\}$ of spatial coordinates of the sequential positions on which the vehicle should be.
The executed (or observed) path is recorded as a time series  $\mathcal{V} = \{(x_{V,i},y_{V,i}), i = 1,\dots,n_V\}$ of GPS coordinates.
Typically $n_V$ is much larger than $n_R$. In Figure~\ref{fig:hci-player} we have $n_R=2925$ and $n_V=4744$.
The points in $\mathcal{R}$ are automatically sampled by {\em DeepRuler} from the analytical equation of the target curve and transmitted to the controller of the vehicle.  
In {\it DeepRuler} for each of the four phases in a run the reference points to be followed are transmitted as a block to the controller just before the start of the phase.

One could argue that this implies that {\em DeepRuler} is likely to compare different controllers rather than different vehicles. 
Instead its applicability is much more general. 
Indeed, we are interested in  the simultaneous evaluation of the hardware and the software of the vehicle: mechanics, manoeuvrability, control. 
A typical situation occurs when for the execution of a path-following task the choice is between two possible UMVs.
One would choose the vehicle for which most often the maximal distance between reference and target paths is smallest. 
Often one needs to consider geometrical performance indices based on $\mathcal{R}$ and $\mathcal{V}$ but also non geometrical performance indices or a combination of them are required at times. 
Some of these are discussed in Section~\ref{sec:performanceIndices}.

%---------------------------------------------------------------------
%--------------------------------------------

\section{Performance indices}
\label{sec:performanceIndices}
We present an overview of quantitative indices currently adopted  for measuring the performance in a path-following execution at sea. 
As mentioned in Section~\ref{sec:experimentDefinition}, we want to characterize the robot capabilities in terms of {\em absolute} performance. 
With respect to classical indices in naval engineering, e.g. tactical diameter and first overshoot angle on zig-zag manoeuvers~\cite{Belenky2006}, we do not involve the vessel size. 
Indices similar to those below can be defined by following such naval indicators, e.g. dividing by the length of the vehicle.
%but their interpretation will have to be different from the one adopted here.

Mainly we consider two types of indices: measures of how far the actually performed path is from the target path, and measures of efficiency, such as financial costs and stress of the mechanics~\cite{Saggini2014performance, caharija2015comparison, nadj2015navigation}. The reference and the observed path are given by $\mathcal{R}$ and $\mathcal{V}$, as in Section \ref{sec:experimentDefinition}. The points in $\mathcal{R}$ can be given as an ordered sequence of  points sampled from an analytical curve $\gamma$, usually assumed unknown  during the post-processing phase.
%{\Ele~\cite{} (in sospeso perche' non sappiamo un buon riferimento biblio e una buona motivazione)}. 
Specifically, each run of the experiment is characterized by its own $\gamma$.
%and its own $\mathcal{R}$ and $\mathcal{V}$.  
When considering forth and back executions, time series of the vehicle positions $\mathcal V_F$ and $\mathcal V_B$ are associated to the given sets $\mathcal R_F$ and $\mathcal R_B$. 
The average of the performance indices evaluated on $\mathcal V_F$ and $\mathcal V_B$ is considered for each pair of back and forth runs. 
%are associated to $x$. We want to model this mean values $y$ for each univariate performance index and seek a model from $x$ to $y$.
One could choose not to take the average performance.
We adopt it following feedback from simulated experiments which show that there is no statistically significant difference of performances on forward and backward paths~\cite{Sorbara2015}.
Also this is confirmed for three out of four field experiments executed under favourable external conditions.  
If the performance indices for the forward and backward paths are very different, then one could choose not to aggregate them. 
But, we stress, aggregation is strongly advocated by the UMV experts in our team.

At the moment, for each run of an experiment {\em DeepRuler} gives an output  $\mathcal{V}$ and the vehicle depths which presently we do not use. 
The number and type of outputs can be easily extended according to the available sensors on the unmanned vehicle to be tested.
A typical telemetry for UMVs includes roll, pitch and yaw angles, robot velocity. They could be exploited in the definition of other performance indices. 
For guaranteeing large applicability, it is suggested to consider quantities that are commonly available in the telemetries, or, at least, to consider quantities that characterize classes of vehicles, e.g. those with a rudder.

Finally, performance indices are required to be easy to compute because we would like to consult them online. %the vehicle to compute them online. 
They are useful for spotting difficulties during the execution of the experiments allowing prompt intervention, for better data quality, for reducing post-processing time, for defining adaptive designs and for allowing adaptations to path-tracking, among other reasons.
%While reviewing the main performance criteria currently adopted, we also present the structure of the data. 
The first set performance indices we consider is apt to measure geometric accuracy. 

\begin{itemize} 
\item {\bf The area index}, $D_A$ is the mean area between $\mathcal{R}$ and $\mathcal{V}$.
% by the total length of $\mathcal{R}$. It measures the mean distance between the two paths. 
For its estimation the paths are regarded as two finite chains of straight line segments, and the areas of non-self-intersecting and consecutive polygons are evaluated with the Gauss area formula. Specifically, let $P_1,\dots,P_{n_P}$ be the polygons identified by the two paths, then 
\begin{equation*}
D_A = \frac{\sum_{j=1}^{n_P} A_{P_j}}{L},
\end{equation*}
where $L$ is the length of the reference path and $A_{P_j}$ is the area of $P_j$. 
If $P_j$ has $n_j$ sides and vertices $(x_i,y_i)$ for $i=1,\dots,n_j$, then its area is 
\begin{equation*}
A_{P_j} = \frac{1}{2}\abs{\sum_{i=1}^{n_j} (x_i y_{i+1}-x_{i+1}y_i)} 
\end{equation*}
where $(x_{n_j+1},y_{n_j+1})=(x_1,y_1)$.
The length $L$ is estimated as the sum of the Euclidean distances between consecutive points. 
\item {\bf The Hausdorff distance}, $D_H$ is defined as 
\begin{equation}%\label{def:hausdorff}
  D_H = \max\{ d_{H}(\mathcal{V},\mathcal{R}), d_{H}(\mathcal{R},\mathcal{V})\},
\end{equation}
where $d_{H}(\mathcal{V},\mathcal{R})$ is the directed Hausdorff distance from $\mathcal{V}$ to  $\mathcal{R}$, 
\begin{equation*}
 d_{H}(\mathcal{V},\mathcal{R}) = \max_{v \in \mathcal{V}} \{\min_{r \in \mathcal{R}} d(v,r)\}
\end{equation*}
and $d$ is the Euclidean distance. It measures 
the maximum of the distances from a point in a set to the closest point in the other set. Thus it is an indication of the maximum distance between $\mathcal{R}$ and $\mathcal{V}$.
\item {\bf The cross-track error}, $XTE$ is the normal component of the distance between the actual vehicle position and the desired position on the target path, with respect to the Frenet-Serret frame on the desired point. It depends on the implementation of the control algorithm and its interpretation is not always straightforward, even if clearly related to the geometric accuracy. Consider for instance a vehicle in proximity of a reference circle. Even if the geometric distance to the closest point on the circle can be objectively evaluated, the $XTE$ is related to the distance between the vehicle position and the position it should reach on the circle. This is a typical choice to avoid overshoot effects. Besides this, in the perspective of sharing data and comparing results from different control schemes, the interpretation of $XTE$ can be even more misleading. However, it is worth noting that in the majority of tested paths (straight lines or combination of line segments) this seems less relevant, and $XTE$ remains the most commonly adopted index because it is provided as a feedback output of all control algorithms. The mean $XTE$, together with the standard deviation is used often~\cite{caharija2015comparison, nadj2015navigation, Saggini2014performance}. 
%In conclusion, even though these measures are frequently presented as valid for the path-following accuracy, their interpretation as ``capability of the  seems more appropriate.
\item {\bf The crossing cell index}, $P_{CA}$ counts the percentage of points in $\mathcal{V}$ that are classified as close to $\mathcal{R}$ more than a given tolerance. This classification is defined by two steps: 1) the reference path is approximated by a polynomial curve and 2) for each point in  $\mathcal{V}$ the {\em Crossing cell algorithm} is applied for deciding whether the zero-locus of the given polynomial intersects the neighbourhood of the  point~\cite{saggini2014assessing,torrentebeltrametti2014}.  
\end{itemize}
A theoretically desirable performance index should be a  function of the exact distance between a point in $\mathcal{V}$ and the target path, that is the zero set of $\gamma(x,y)=0$.  This could be computed by solving a constrained minimization problem e.g. using Lagrange multipliers. For many $\gamma$ this computation is not sufficiently efficient and speedy for begin performed during path-following. Furthermore the numerical computation of the zero set of $\gamma$ might be cumbersome, time demanding and not always possible. Considering that $P_{CA}$ requires difficult analytic approximation, the previous, theoretically less appealing, indices are preferred. 

The most popular performance indices used for characterizing the efficiency are as follows. 
\begin{itemize}
\item {\bf The XTE decreasing rate} is 
defined as the maximum (or mean or ...) value of the series of the differences of consecutive XTE samples. It measures the promptness of the system to react when a new reference is given. Thus it is investigated only during transient phases, when it is clearly decoupled from steady state phase~\cite{Saggini2014performance}.
\item {\bf The rudder stress} is a measure of the stress on the mechanics of the vehicle during manoeuvering. The mean and/or the maximum values of the absolute values of the rudder angle are usually computed~\cite{caharija2015comparison, Saggini2014performance}.
\item {\bf The thrusters' energy consumption} measures the mean action on the thrusters. It is obtained by averaging the commanded thrust force multiplied by the distance between consecutive positions~\cite{Saggini2015evaluation}.
\end{itemize}

In some applications it might be of interest to consider the above  indices in relationship to some characteristics of the path, e.g. by taking into account some notion of ``difficulty'' in executing a path-following task. For relevant geometric definitions of this difficulty see e.g.~\cite{lekkas2014guidance}.   
Experience from field tests %in marine mobile robotics 
suggests to take into account the path curvature, e.g. mean and/or maximum value, or the presence of turns, their number and closeness. 
Finally a current interest is on the definition of compound performance indices. Since it is uncomfortable to consult a long list of indices, a practical solution could be to summarise all the indicators in a single and easily interpretable value: to do this, joint interpretation of geometric and other types of indices is fundamental and needs to be further investigated.

%---------------------------------------------------------------------
\section{Approximation models for a family of target paths}
\label{sec:dace}

In this section we present the approximation model for the performance indices introduced in Section~\ref{sec:performanceIndices}. We consider experiments for which each run is from the same family of curves expressed via few parameters because at the moment the typical field trials are repetitions of executions of the same target path. The design problem is on the parameters.  
Thus with an experiment we test a class of curves $\gamma_x$ with $x \in {\mathcal X} \subset \mathbb{R}^q$, for example $q=2$ in a sine wave. Both implicit as in~\cite{saggini2014assessing} and explicit equations for the path curves can be adopted. 
We want to model univariate mean performance indices $y$ and seek a model from $x$ to $y$.

As in  an experiment there are $n$ runs, the design is given by $n$ values, say $S = [s_1, \dots,s_n]^T$ with each design point $s_i$ in the design space ${\mathcal X}$ and $A^T$ indicates transpose of $A$. 
Responses are  $Y = [y_1, \dots,y_n]^T \in \mathbb R^n$, with $y_i=y(s_i)$ and $s_i$ gives the path $\gamma_{s_i}$. 
The choice of the input set $S$ at which data should be collected is addressed in a semi-automatic fashion and is described in Section~\ref{sec:doe},  
while the choice of ${\mathcal X}$ relies on the family of curves 
best suited according to the experts for testing the vehicle. Next we list some  examples:
\begin{itemize}
\item {\bf Circles} are a natural choice for exploring the minimum turning radius. The execution of path-following 
is clearly invariant to translations and the paths are fully described by the radius parameter. Hence, $q=1$ and ${\mathcal X} = \mathbb{R}^{+}$.
\item {\bf Sine waves} are used for testing the vehicle on different curvatures using the same run. A sine wave is  determined by an amplitude and a period. %, or equivalently the number of waves in $W$. 
We adopt the number of hemi-periods as the second parameter because we want complete hemi-periods within $W$.
This technical requirement has positive repercussions on achieving repeatibility, as mentioned in Section~\ref{sec:experimentDefinition}, and on the definition of performance indices and their use for performance comparisons.
We have $q=2$ and ${\mathcal X} \subset \mathbb{R}^{+} \times \{0,1,2,\ldots\}$. 
\item {\bf Square waves} are suitable paths for testing  steady state on the line and the promptness of response to follow new inputs. Similarly to the sinusoidal case, we have $q=2$ and ${\mathcal X} \subset \mathbb{R}^{+} \times 
\{0,1,2,\ldots\}$. 
\end{itemize}

Following standard theory~\cite{lophaven2002dace, sacks1989design, rasmussen2006gaussian}, we adopt a model that treats the deterministic response $y(x)\in \mathbb{R}$ as a realization of a stochastic process with a linear deterministic component. 
For  $x \in \mathbb{R}^q$ we write
\begin{equation}\label{eq:Kmodel}
{Y}(x) = \sum_{j=1}^{p}\beta_jf_j(x) + Z(x),
\end{equation}
where $Z(\cdot)$ is a random process with zero mean and covariance between $Z(w)$ and $Z(x)$ given by 
\begin{equation*}
E[Z(w)Z(x)] = \sigma^2\mathcal{R}(\theta, w,x),
\qquad x,w\in \mathbb{R}^q .
\end{equation*}
Here $\sigma^2$ is the process variance and $\mathcal{R}(\theta,w,x)$ is the correlation between $x$ and $w$. We use the following notation
\begin{enumerate}
\item $f(x) = [f_1(x),\dots,f_p(x)]^T$  for the $p$ regression functions in~(\ref{eq:Kmodel}), 
\item $F = [f_j(s_i)]_{i=1,\dots, n,j=1,\dots,p}$ 
for the $n \times p$ design matrix, %[f(s_1),\dots,f(s_m)]^T \quad \text{for the $m \times p$ design matrix},
\item $R =  %[R_{ij}]_{i,j=1,\dots, m}= 
[\mathcal{R}(\theta,s_i,s_j)]_{i,j=1,\dots, n}$
for the correlations at design points, and
\item $r(x) =  [R(\theta,s_1,x),\dots,R(\theta,s_n,x)]^T$ for the vector of correlations between the design sites and an untried input $x$.
\end{enumerate} 
The best linear unbiased predictor (BLUP) for $y(x)$ is
\begin{equation}\label{eq:BLUP}
\hat{y}(x) = f^{T}(x)\hat{\beta} + r^{T}(x)R^{-1}(Y-F\hat{\beta}),
\end{equation}
where $\hat{\beta}=(F^{T}R^{-1}F)^{-1}F^{T}R^{-1}Y$ is the generalized least-square estimate of $\beta$. The mean square error $\MSE[\hat{y}(x)] $ is equal to
\begin{equation}\label{eq:MSE}
\sigma^2\left[1-(f(x)^Tr(x)^T) \left( \begin{array}{cc}
0 & F^T \\
F & R \end{array} \right)^{-1} \left( \begin{array}{c}
f(x)  \\
r(x)  \end{array} \right) \right].
\end{equation}
For details on how to derive~(\ref{eq:BLUP}) and~(\ref{eq:MSE}) see e.g.~\cite{sacks1989design}.

Typically the experimenter chooses  $f(x)$ and the correlation function $\mathcal{R}(\theta,w,x)$, while $\sigma^2$ is  given by maximum likelihood 
estimation.
A low-order
polynomial $f$ is often appropriate, so that typical choices are the constant, linear and quadratic models. 

Even if it seems a trivial assumption at first sight, it is quite common in engineering applications to choose the constant trend model. This is a way to rely upon the correlation model to ``pull'' the response surface through the observed data by quantifying the correlation of nearby points~\cite{martin2004use}. 

The correlation model is often chosen to be the product of stationary one-dimensional correlations
\begin{equation*}
\mathcal{R}(\theta,w,x) = \prod_{j=1}^{q}\mathcal{R}_j(\theta,w_j-x_j), \qquad 
w,x\in \mathbb R^q.
\end{equation*}
The choice of $\mathcal{R}_j$ should depend on the knowledge of the underlying phenomenon. Some hints are given for instance in~\cite{sacks1989design}, together with classical choices, like the exponential, gaussian, linear, cubic and spline. For the simulation study in Section~\ref{sec:sim} we adopt the gaussian model 
$\mathcal{R}_j(\theta,w_j-x_j)= e^{-\theta_j(w_j-x_j)^2}$.
A number of toolboxes and packages for computing kriging approximations are freely available in softwares like R, Matlab, Scilab and Phyton.
The accessibility of efficient and accurate software has favoured the use of kriging in engineering applications in the last ten years~\cite{martin2004use}.
In Section~\ref{sec:sim} we use the DACE kriging toolbox of Matlab~\cite{lophaven2002dace}. 

%---------------------------------------------------------------------
\section{Adaptive design of experiments}
\label{sec:doe}
For a family of target paths % as in Section~\ref{sec:dace} 
and a performance index, 
the design problem consists in selecting some paths the vehicle should follow either in order to assess the best/worst performance achievable or in order to reconstruct the performance function.
For both design problems, we adopt a two-step procedure  and the performance function is the response function modeled in Equation~(\ref{eq:Kmodel}). 
The design space $\mathcal{X}$ is usually given by 
a subset of $\mathbb R^q$  
and the reconstruction space is (the smallest) convex set 
containing $\mathcal{X}$. 
For sine-waves $\mathcal{X}$ is given by parallel equal-length segments in $\mathbb R^2$, one segment for each allowed hemi-period, together with the point $(0,0)$ representing the straight line. 
At the moment, the experimenter is in charge of selecting $\mathcal{X}$ by taking into account the manoeuvering capabilities of the vehicle. The GPS sensor accuracy, from which $\mathcal V$ depends, is in this context not a relevant parameter since it is typically much higher than the manoeuvre accuracy. Eventually $\mathcal{X}$ might be defined automatically out of a small number of pilot tests and the vehicle  characteristic parameters.

For path-following, some pilot runs, almost always straight lines, 
are executed prior to conducting the experimental procedure.
Thus we include a back and forth straight line run in all the  experiments. 
These runs are especially important in real scenarios, because they provide preliminary information on the measurement system and give a rough idea of experimental errors. 
They are used for verifying the correct operation of sensors and communications systems, and for setting-up the robot control system parameters, if required.

%--------------------------------------------------------------

\subsection{Worst Performance Design} %/Optimizing a performance function
\label{subsec:WP}

We consider a performance index $y$ for which high values correspond to bad executions of path-following %, as for the indices in Section~\ref{sec:performanceIndices},
and we are interested in high value local maxima.
Let $\mathcal{X} \subset \mathbb{R}^q$ be the design region, and $n_1$ and $n_2$ be the number of design points of the two steps, respectively. 
The choice of $n_1$ relies upon the total time available, the design space $\mathcal X$ to be investigated and engineering requirements and know-how. 
In the first step, a space filling design, such as a latin hypercube design (LHS) or a one generator lattice, is recommended even if prior knowledge would suggest to concentrate the search of local maxima in a restricted portion of $\mathcal X$. Indeed in Section~\ref{sec:sim} we present an example for which the area index has maxima in an unexpected subset of $\mathcal X$. 

The first $n_1$ design points $S_1 = \{s_1,\dots,s_{n_1}\} \subset \mathcal{X} $ always includes the straight line: if, after its execution back and forth, there is no need to change the experimental set-up, e.g. the set-up of the controller, $R_1$, $W$ or $\mathcal X$, we do not replicate it, in order to save time and energy. If, instead, the experimental set-up has to be adjusted, then only the last replicate of the straight line is considered for the subsequent analysis.

The $S_1$ experiment is executed and the set of responses $Y_1 = \{y_1,\dots,y_{n_1}\}$ is computed. 
The maxima values can be pointed out in different ways: 
a) by the index $i_M$ for which $y_{i_M} = \argmax Y_1$, or 
b) by all $i$ for which $y_i > \epsilon_{MAX}$ where $\epsilon_{MAX}$ is a given threshold, or
%is given, then we take all the $i$ such that $y_i > \epsilon_{MAX}$ or 
c) by all $i$ for which $\abs{y_{i_M}-y_i}<\epsilon_{DIFF}$ where $\epsilon_{DIFF}$ is a given tolerance on the difference. 
%if a tolerance on the difference $\epsilon_{DIFF}$ is given, we take $\overline{i}$ and all the $i$ for which $\abs{y_{\overline{i}}-y_i}<\epsilon_{DIFF}$. 
It is expected that some maxima are located on the subset of the boundary of $\mathcal X$ which encodes the seemingly most difficult to follow paths, 
%in the parametrization we have chosen to identify path 
although this is partly confuted by some tests shown later. But what is most concerning in practice and should be investigated, is the presence of internal maxima in $\mathcal X$.
Next, other $n_2$ design points $S_2 = \{s_{n_1+1},\dots,s_{n_1+n_2}\}$ are selected in proximity of the identified maxima local in the first step and responses $Y_2 = \{y_{n_1+1},\dots,y_{n_1+n_2}\}$ are computed. The number $n_2$ is again chosen according to the remaining time available and the other usual constraints. For example, a central composite design (CCD) can be allocated to each  of the previous identified maxima. If some of the CCD points fall outside $\mathcal X$ or have been already executed, then they are redistributed in $\mathcal X$ randomly. 

A kriging model $\hat{y}^{WP}$ is computed for inputs $S_1 \cup S_2$ and responses $Y_1 \cup Y_2$. Finally, the model is evaluated on a thick grid of points $S^{\ast} \supset S_1 \cup S_2$ and  the final estimation of the maximum is given by $\hat{y}_{\text{MAX}}^{WP} = \max_{s \in S^{\ast}}\hat{y}^{WP}(s)$ or other important local maxima can be estimated. %Furthermore, let $y$ be the performance function to be optimized. %Therefore, the goal is to estimate the the maximum of $y$. %The classical theory on response surfaces can be adapted to the Kriging approximation: in fact, even if  traditional search methods (like steepest ascent and grid search) cannot be properly applied (since the model $\hat{y}$ strongly depends on the correlation structure), the study of the maximum is naturally local, and central composite designs are suitable for it.
%A major interest for moving towards a better knowledge or characterization of a marine robotic systems is to investigate about its maneuvering capabilities, and thus looking for the maximum of $\eta$. 
% In the following box the steps of the proposed sequential strategy are summarized. 
The above procedure generalises for the minimum value search or by including more sequential steps, straightforwardly.

Various performance indices $Y_{FF}$ can be observed on a dense full factorial grid $S_{FF} \subset \mathcal{X}$ and a model $\hat{y}^{FF}$ and a maximum estimated value $\hat{y}_{\text{MAX}}^{FF}= \max_{s \in S^{\ast}}\hat{y}^{FF}(s)$ can be estimated over a thicker grid $ S^{\ast} \supset S_{FF}$ for each index $y$ in $Y_{FF}$.
Then we performed the above two-step procedure with $S_1 \cup S_2\subset S_{FF}$, obtaining a model $\hat{y}^{WP}$ and a maximum estimated value $\hat{y}_{\text{MAX}}^{WP}= \max_{s \in S_{FF}}\hat{y}^{WP}(s)$, whose goodness %of estimation of the maximum of $y$ through the model $\hat{y}^{WP}$ with respect to $\hat{y}^{FF}$ can be measured by
is assessed by $ABS=\abs{\hat{y}_{\text{MAX}}^{WP} -\hat{y}_{\text{MAX}}^{FF}}$.

%-------------------------------------------------------
\subsection{Best Prediction Design} %Predict the performance function at untried inputs/Reconstruct the performance function over a portion of the design space}
\label{subsec:BP}
Next we adapt the two-step procedure of Subsection~\ref{subsec:WP} to the reconstruction of the performance function in a desired region $S^\ast$. 
The first step with $n_1$ points is as above and the responses $Y_1 = \{y_1,\dots,y_{n_1}\}$ are associated to the design points $S_1 = \{s_1,\dots,s_{n_1}\}$. 
%With the  notations and set-up of Subsection~\ref{subsec:WP}
%We consider notations and the first experimental step as in Subsection~\ref{subsec:WP}, and describe how to continue the experimental procedure when the aim is to reconstruct the performance function in a desired region of the design space.  
%Assume that responses $Y_1 = \{y_1,\dots,y_{n_1}\}$ have been associated to the first set of design points $S_1 = \{s_1,\dots,s_{n_1}\}$.
Then, a kriging model $\tilde{y}^{BP}$ is computed and the $\MSE(\tilde{y}^{BP})$ is evaluated on a thick grid of points $S^{\ast} \supset S_1$. 
We select  points in $S_1$ with high $\MSE$
%choose among the design points those corresponding to high values for the $\MSE$ 
and choose the next $n_2$ design points in their proximity. A kriging model $\hat{y}^{BP}$ is computed for inputs $S_1 \cup S_2$ and responses $Y_1 \cup Y_2$
and it is evaluated on the grid $S^{\ast} \supset S_1 \cup S_2$.

As in Subsection~\ref{subsec:WP}, to validate the procedure, we consider a kriging model $\hat{y}^{FF}$  for a dense full factorial grid $S_{FF}$ with $n_{FF}$ points and corresponding responses $Y_{FF}$. Indications of accuracy in the reconstruction of $y$ through the model $\hat{y}^{BP}$ can be measured by 
\begin{equation*}
\max ABS_{error} = \max_{s \in S_{FF}} \abs{Y_{FF}(s)-\hat{y}^{BP}(s)}
\end{equation*}
\begin{equation*}
\mean ABS_{error} = \sum_{s \in S_{FF}} \frac{\abs{Y_{FF}(s)-\hat{y}^{BP}(s)}}{n_{FF}}  .  
\end{equation*}
Since the kriging model is a perfect interpolator, the values $\abs{Y_{FF}(s)-\hat{y}^{BP}(s)}$ for points $s \in S_1 \cup S_2$ are zero.

%---------------------------------------------------------------------
%--------------------------------------------------
\section{A simulation study}
\label{sec:sim}
Various tests were conducted with the Charlie USV hardware-in-the-loop simulator~\cite{caccia2009charlie} and \emph{DeepRuler} for proving the reliability of \emph{DeepRuler} and of the above procedure for executing and replicating experiments. 
Path-following experiments were performed for the family of sinusoidal paths 
\begin{equation}\label{eq:sin}
\gamma_{(x_1,x_2)}(w) = x_1 \sin\left(\frac{\pi x_2}{W}w\right), \quad w \in [0,W], 
\end{equation}
with $[x_1,x_2]^T \in {\mathcal X} \subset \mathbb{R}^{+} \times \{0,1,2,\ldots\}$. 
Initial results for simulated test and on sea trials are reported in~\cite{Sorbara2015}.

%and a field campaign has been promised for next spring. 
The simulator is part of the custom Charlie USV architecture, is developed in C++ for standard Linux O.S. distributions and it allows a very precise simulation of the vehicle dynamic and kinematic motion evolution. Each point $(x_{V,i},y_{V,i}) \in \mathcal{V}$ is affected by an additive error $(\varepsilon_x,\varepsilon_y)$ with $\varepsilon_x$ and $\varepsilon_y \sim Unif(-a,a)$ independent and independence holds also across the $(x_1,x_2)$ positions. A more accurate and realistic model for the measurement process could be considered by including models for the weather and the external conditions or other probability distributions for the added errors. This would overload the system unnecessarily and, at least for the moment, we assume that any external influence is embedded in the correlation structure of the model in Equation~(\ref{eq:Kmodel}) and in the $\varepsilon_x$ and $\varepsilon_y$. 

\begin{figure}[!ht]
 \centering
 \begin{minipage}[c]{0.48\textwidth}
 \centering
 \includegraphics[width=\textwidth]{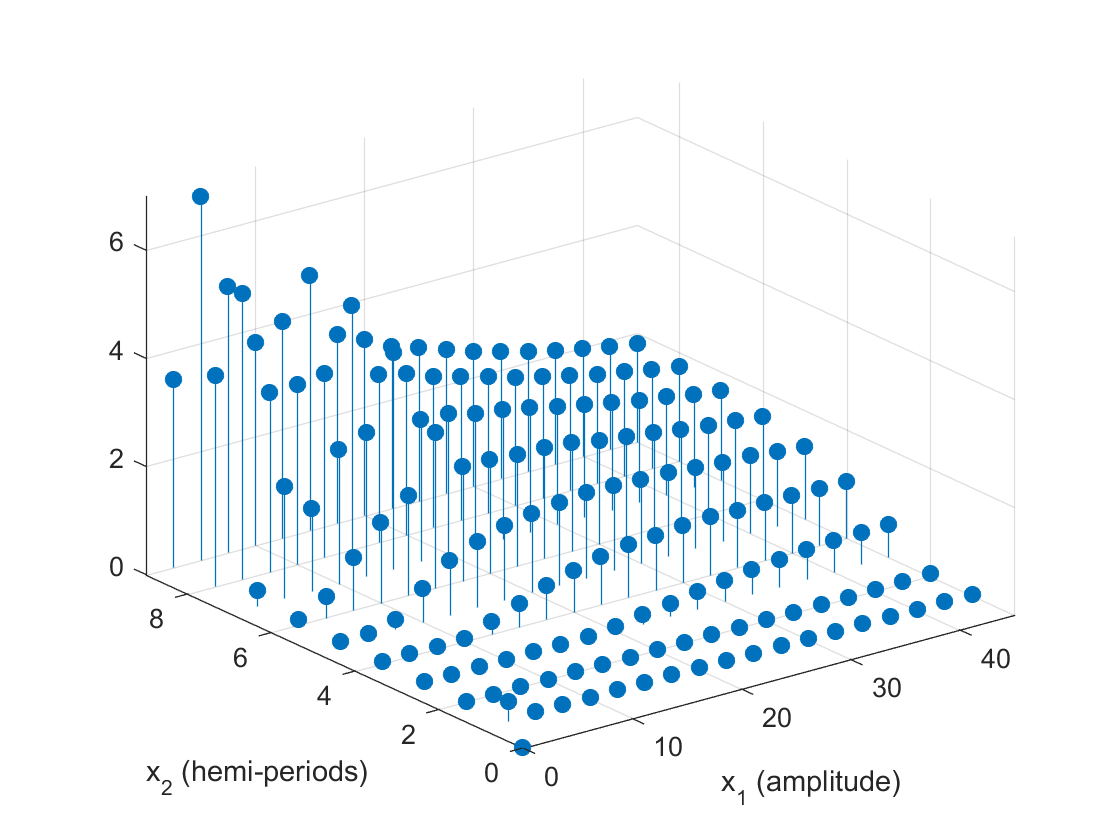}
 \caption{Performance index $D_A$ on $\{(0,0)\} \cup\{2.5i:i=1,\dots,18\} \times \{i:i=1,\dots,9\}$.}\label{fig:areaFF}
 \end{minipage}%
\hspace{0.3cm}
\begin{minipage}[c]{0.48\textwidth}
 \centering
 \includegraphics[width=\textwidth]{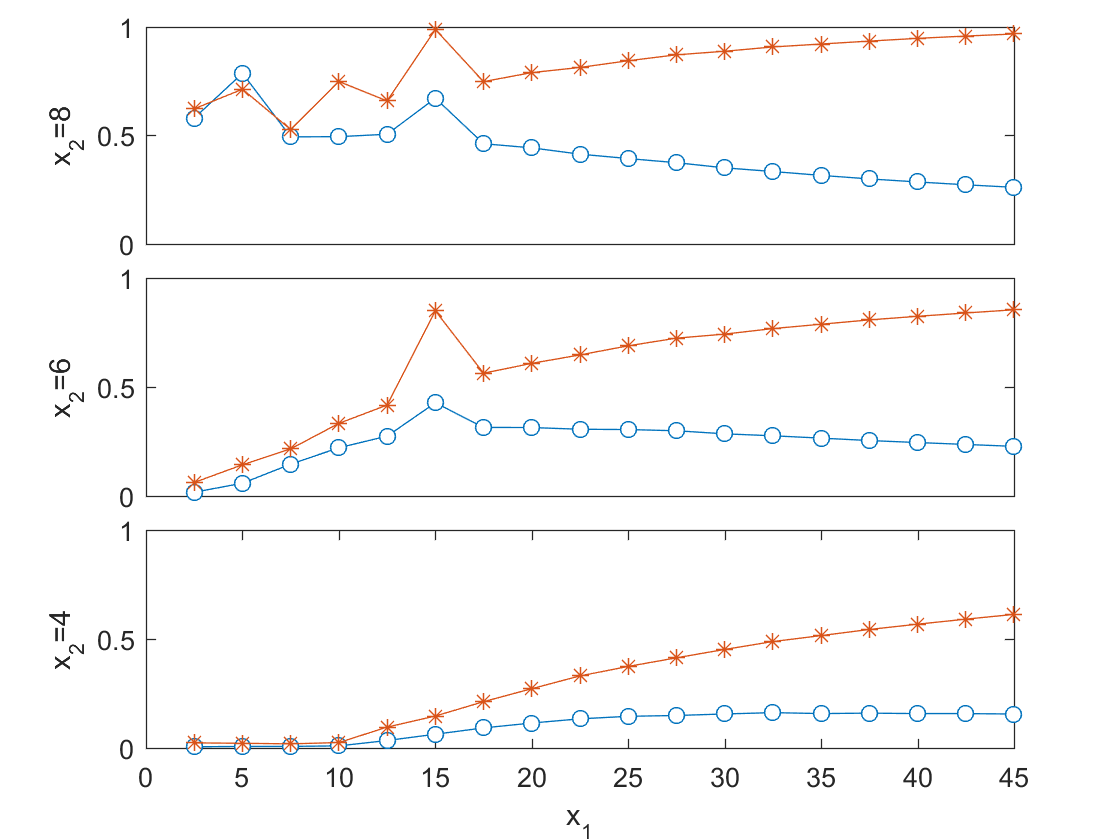}
 \caption{Indices $D_A$ (blue dots) and $D_H$ (red stars) normalized in [$0$,$1$] computed for hemi-periods $4$, $6$ and $8$ and amplitudes in $\{2.5,5,\dots,45\}$.}
 \label{fig:cfr}
\end{minipage}
 \end{figure}
 
%   \begin{figure}[htbp]
%  \centering
%  \begin{minipage}[c]{0.48\textwidth}
%  \centering
%  \includegraphics[width=\textwidth]{krigingAreaFF_poly0.png}
%  \caption{Kriging reconstruction of $D_A$ with a constant regression term and gaussian correlation with $199$ points.}\label{fig:areaFFkriging}
%  \end{minipage}%
%  \hspace{0.3cm}
%  \begin{minipage}[c]{0.48\textwidth}
%  \centering
%  \includegraphics[width=\textwidth]{krigingHausFF_poly0.png}
%  \caption{Kriging reconstruction of $D_H$ with a constant regression term and gaussian correlation with $199$ points.}
%  \label{fig:hausFFkriging}
%  \end{minipage}
%  \end{figure}
 
%----------------------------------------------------------------
\subsection{Preliminary Simulation Tests}
\label{subsec:firstSim}
The tests with the Charlie USV simulator and \emph{DeepRuler} were performed with $W=100$  $m$, $R_2=110$ $m$, $l=25$ $m$, $r=14$ $m$, $a=0.2$ $m$ and uniform noise added to the observed $\mathcal V$ and also to the vehicle heading ($0.2^\circ$ maximum) and speed ($0.1$ $m/s$ maximum).
In the earliest trials ${\mathcal X}=\{(0,0)\} \cup\{5,10,15,20,25\} \times \{1,2,3,4,5,6\}$, then the inputs were extended up to $35$ for the amplitude and $7$ for the hemi-periods, and the last simulated data set was with  ${\mathcal X}=\{(0,0)\} \cup\{5i:i=1,\dots,11\} \times \{i:i=1,\dots,9\}$ and zero measurement noise. 
The zero noise case of the last data set satisfies as much as possible the condition of replicability of results for computer experiments. 
 
All these tests have been instructive and helpful for confirming and supporting most decisions made during the project and reported in the previous sections. Results of paired samples $t$-tests confirmed our prior assumptions about the independence between forward and backward executions and about the independence on the order of execution of the runs within an experiment. %What is more important is the knowledge acquired about the system from those tests: % lesson learned:
Moreover, the values of $D_A$ and $D_H$ suggested to consider a correlation model for the points in ${\mathcal X}$ and thus the kriging model in Section \ref{sec:dace} was chosen.

Our attention was also focused on the response to be attached to the border points 
$\{(x_1,0), x_1 \in \mathbb{R}^+\}$ and 
$\{(0,x_2), x_2=0,1,\dots\}$. They all correspond to the ``straight line'' case like the point $(0,0)$. %, the simplest path to follow and thus with expected small values of the performance indices. 
Even if at first it seemed natural to extend the values $D_A(0,0)$ and $D_H(0,0)$ to the left and bottom borders, a suspicious peak near the point $(5,8)$ visible in Figures~\ref{fig:areaFF} and~\ref{fig:cfr} suggested to obtain further samples for amplitudes smaller than five. % and the corresponding values of $D_A$ and $D_H$. 
The values obtained with ${\mathcal X}=\{(0,0)\} \cup\{0.5i:i=1,\dots,5\} \times \{i:i=1,\dots,9\}$ are not reported here. They confirm the hypothesis of a rapid slope from the peak towards the near border, ultimately supporting the assignment of the values $D_A(0,0)$ and $D_H(0,0)$ to all border points
$(x_1,0)$ and $(0,x_2)$ with $ x_1 \in \mathbb{R}^+$ and $x_2=0,1,\dots$. 
 
\begin{figure}[htbp]
 \centering
 \includegraphics[width=0.48\textwidth]{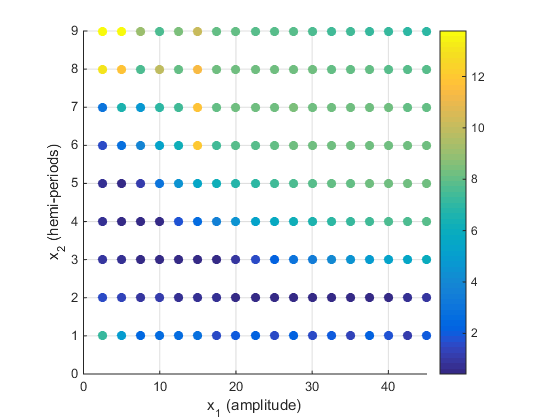}
 \caption{Heatmap of $D_H/\sqrt[3]{k_{MAX}}$, interpreted as an index for evaluating the  geometric accuracy of path-following execution with respect to path complexity.}%  with the Hausdorff distance in the relation to the complexity of the reference path.}
 \label{fig:DHdividebByCubeRootcAbsMax_CSDA}
 \end{figure}

Finally, there was a prior belief among engineers that curves with larger parameter values were more complex and thus harder to follow, providing larger performance indices. Figures~\ref{fig:areaFF} and~\ref{fig:cfr} are counter-examples for $D_A$ and $D_H$. As a measure of complexity of a curve in Equation~(\ref{eq:sin})
one can consider the maximum value of its curvature, given by the ratio between the second derivative in $w$ and the $3/2$-root of one plus the squared first derivative. 
For sine waves this is achieved in $w = -x_2W/2$ and it is 
\begin{equation*}%\label{eq:kMax}
k_{MAX} = x_1\left(\displaystyle{\frac{\pi x_2}{W}}\right)^{2}.
\end{equation*}
%This is  achieved in $w = -x_2/4-W/2$ and it is \[
% k_{MAX} = \frac{2\sqrt{2}x_1\displaystyle{\frac{\pi^2}{x_2^2}}}{\left(1+2\left(x_1\displaystyle{\frac{\pi}{x_2}}\right)^2\right)^{3/2}}.
% \] 
% as suggested by the engineers, we expected a relationship between the maximum curvature of a path and the complexity of execution. %Since the maximum of the absolute value of the curvature of $\sin l$ is achieved in $l=\pi/2$, % in $x = \pi/2 + k \pi$, with $k \in \mathbb{Z}$,
% The absolute value of the maximum curvature of $\gamma$ in Equation (\ref{eq:sin}) is achieved in $w = -x_2/4-W/2$ \todo{(controllare} and it can be computed as 
%\[
%k_{MAX} = \frac{2\sqrt{2}x_1\displaystyle{\frac{\pi^2}{x_2^2}}}{\left(1+2\left(x_1\displaystyle{\frac{\pi}{x_2}}\right)^2\right)^{3/2}}.
%\] 
%Therefore, given the two sinusoids $\tilde{\gamma}:=\gamma_{(\tilde{x}_1,\tilde{x}_2)}$ and $\overline{\gamma}:=\gamma _{(\overline{x}_1,\overline{x}_2)}$ with $\tilde{x}_1 \leq \overline{x}_1$ and $\tilde{x}_2 \leq \overline{x}_2$, the worst performance was expected to be obtained while performing the experiment with $\overline{\gamma}$. 
Figures~\ref{fig:areaFF} and~\ref{fig:cfr} refer to simulated data on ${\mathcal X}=\{(0,0)\} \cup\{2.5i:i=1,\dots,18\} \times \{i:i=1,\dots,9\}$ and zero measurement noise. 
Deviations from the expected behaviour are more striking for $D_A$ and minor but relevant differences are also notable for $D_H$. 
High peaks at hemi-periods larger than five are most interesting. 
They might be due to the joint requirement of turning in less than $14$ $m$, which is the estimated minimum turning radius of Charlie ($x_2>5$), and of following a short small curvature curve (small $x_1$). 
Eventually this could be used to improve estimate of the theoretical $r$ for the vehicle under study. 
The situation does not change when, rather than the performance indices, we plot reasonable functions of them and of the index of complexity of a curve. 
For example $D_H/\sqrt[3]{k_{MAX}}$ in Figure~\ref{fig:DHdividebByCubeRootcAbsMax_CSDA} has high peaks in the center and unusual high values in the lower bottom corner. 
This, to us, confirms the scarce understanding of how to  characterize the behaviour of a USV and of the desired response functions, and it  motivates our work. 
The above conclusion is supported even more when considering the data sets with added noise. 
 
%----------------------------------------------------------------
\subsection{Simulated Data on a Dense Grid}
%A full factorial experiment}
\label{subsec:FF}

\begin{figure}[htbp]
 \centering
 \begin{minipage}[c]{0.48\textwidth}
 \centering
 \includegraphics[width=\textwidth]{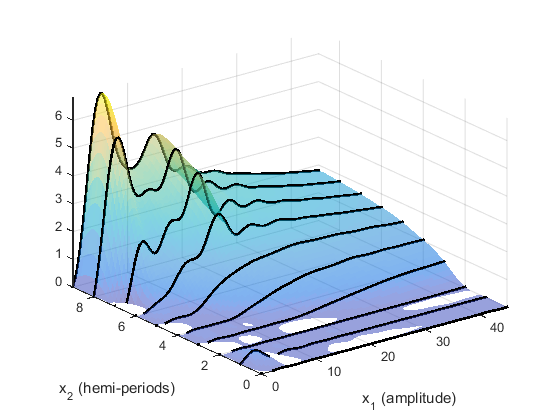}
 \caption{Kriging reconstruction of $D_A$ with a constant regression term and gaussian correlation with $199$ points.}\label{fig:areaFFkriging}
 \end{minipage}%
 \hspace{0.3cm}
 \begin{minipage}[c]{0.48\textwidth}
 \centering
 \includegraphics[width=\textwidth]{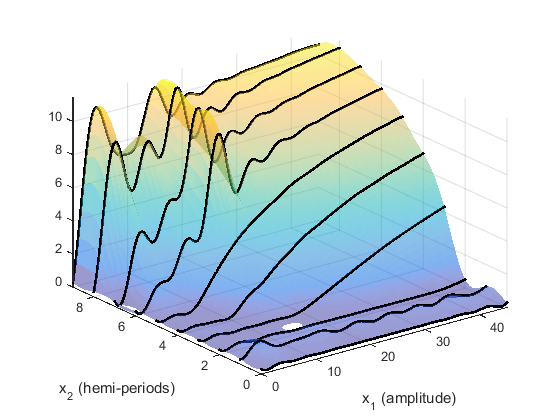}
 \caption{Kriging reconstruction of $D_H$ with a constant regression term and gaussian correlation with $199$ points.}
 \label{fig:hausFFkriging}
 \end{minipage}
 \end{figure}

Using the default input parameters in~\cite{lophaven2002dace}
we computed different kriging models for $D_A$ and $D_H$ and with constant, linear and quadratic regression polynomials. These models %for the Kriging predictor 
and the corresponding mean square errors were evaluated on a regular $100 \times 100$ grid of sample points that contained $\mathcal{X}$. For both $D_A$ and $D_H$, the constant model is preferable because it is the simplest and with low mean ($0.0038$ for $D_A$, $0.0267$ for $D_H$) and with small maximum MSE of the predictor ($0.0173$ for $D_A$, $0.1218$ for $D_H$). The reconstructions obtained are shown in Figures~\ref{fig:areaFFkriging} and~\ref{fig:hausFFkriging}.  
The correlation structure is as in Section~\ref{sec:dace} with $\sigma^2 = 0.9220, \theta_1 = 20, \theta_2 = 3.5355$ for $D_A$ and $\sigma^2 = 6.4812, \theta_1 = 20, \theta_2 =  3.5355$ for $D_H$.
White areas in Figures~\ref{fig:areaFFkriging} and~\ref{fig:hausFFkriging} correspond to negative prediction of $D_A$ and $D_H$. 
These negative values are close enough to zero that we can assume they are due to the expected variability associated to the chosen prediction model. 
%Safely the model can be twicked to take zero value on the white areas. 

\subsection{Adaptive Design for Optimization}\label{subsec:simWP}
Following the algorithm in Subsection~\ref{subsec:WP}, for the first step we randomly took a LHS with $n_1 = 10$ points whose coordinates are in the first two rows on Table~\ref{tab:designPoints}.
For the second step we took a CCD with $n_2=8$ and center on the maximum  of the  performance index observed in the first step. Only two points of the CCD were outside the admissible region or already executed and they were reallocated in $(45,7)$ and $(35,8)$ at random. 
The CCD coordinates are given in the third and fourth rows on Table~\ref{tab:designPoints}.
In the choice of $n_1$ and $n_2$ we considered the feasibility of the two-step procedure on real tests; for instance, the sea test campaign of $10$ runs conducted in~\cite{Sorbara2015} lasted about $2$ hours. 
We started with the same LHS for reconstructing $D_A$ and $D_H$.
Fortuitously also the same points were selected in the second step. 

%  \begin{figure}[htbp]
%  \centering
%  \includegraphics[width=0.48\textwidth]{lhsCCmaxArea_correct.png}
%  \caption{Design points for $D_A$ and $D_H$ used for  Figures~\ref{fig:lhsCCmaxArea_poly1_correct} and~\ref{fig:lhsCCmaxHaus_poly2_correct}, respectively. The dots are for the LHS points, the blue circle is for the centre of the CCD, while stars are the design points of the second step.}\label{fig:lhsCCmax_correct}
%  \end{figure}

\begin{table}[!t]
\renewcommand{\arraystretch}{1.2}
\caption{Coordinates of design points for the analysis in Sections~\ref{subsec:simWP} and~\ref{subsec:simBP}.}
\label{tab:designPoints}
\begin{tabular}{l|cccccccccc}
%\begin{tabular}{c|l| C{0.15cm} C{0.15cm} C{0.15cm} C{0.15cm} C{0.15cm} C{0.15cm} C{0.15cm} C{0.15cm} C{0.15cm} C{0.15cm}}
\multicolumn{11}{c}{First step design is a LHS} \\
\hline
$x_1$ & 0 &5 &10 &15 &20 &25 &30 &35 &40 &45\\
$x_2$ & 0 &5 &3 &8 &7 &2 &6 &1 &9 &4 \\
\hline
\hline
\multicolumn{11}{c}{Second step designs} \\
\hline
 $x_1$ &5 &10 &10 &15 &20 &25 &35 &45\\
 $x_2$ &8 &7 &9 &8 &9 &8 &8 &7\\
\hline
$x_1$ &5 &10 &10 &20 &35 &25 &45 &45\\
$x_2$ &1 &7 &9 &9 &3 &8 &7 &8\\
\hline
\hline
$x_1$ &2.5 &2.5 &5 &7.5 &12.5 &20 &35 &35 &45\\
$x_2$ &7 &9 &1 &8 &9 &9 &3 &8 &7\\
\hline
\end{tabular}
\end{table}

\begin{figure}[htbp]
 \centering
 \begin{minipage}[c]{0.48\textwidth}
 \centering
 \includegraphics[width=\textwidth]{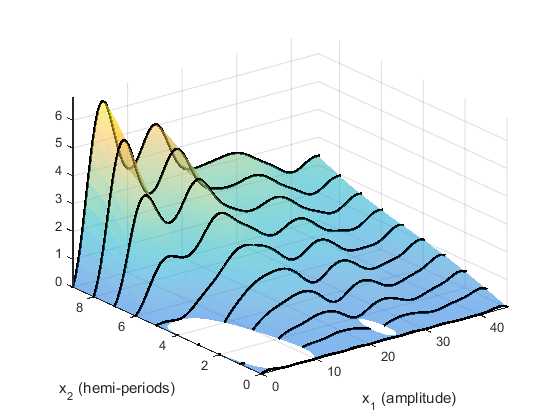}
 \caption{Kriging optimization of $D_A$ with a linear regression term and gaussian correlation with $18$ points.}\label{fig:lhsCCmaxArea_poly1_correct}
 \end{minipage}%
 \hspace{0.3cm}
 \begin{minipage}[c]{0.48\textwidth}
 \centering
 \includegraphics[width=\textwidth]{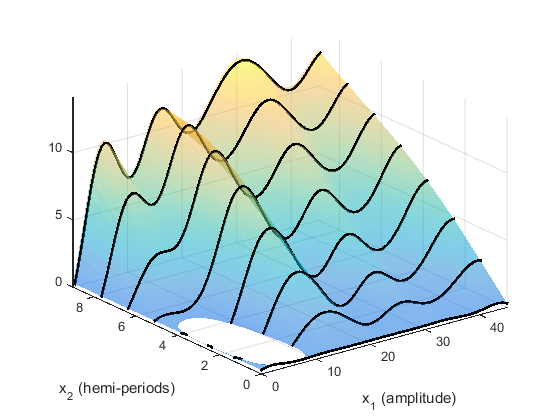}
 \caption{Kriging optimization of $D_H$ with a quadratic regression term and gaussian correlation with $18$ points.}
 \label{fig:lhsCCmaxHaus_poly2_correct}
 \end{minipage}
 \end{figure}

Reconstruction of the performance indices are shown in Figures~\ref{fig:lhsCCmaxArea_poly1_correct} and~\ref{fig:lhsCCmaxHaus_poly2_correct}. 
The linear and quadratic models are preferable for $D_A$ and $D_H$, respectively. %, being those with lower mean %($0.2483$ for $D_A$ and $0.6935$ for $D_H$) and maximum ($1.0019$ for $D_A$ and $2.8733$ for $D_H$ ) MSE. 
For $D_A$ the covariance structure parameters are $\sigma^2= 3.2471$, $\theta_1 = 10$ and  $\theta_2 = 0.3125$, while for $D_H$ they are $\sigma^2=  8.8507$, $\theta_1 = 10$ and  $\theta_2 = 0.3125$. 
Concerning the goodness of estimation of the maximum value, the ABS error in Subsection~\ref{subsec:WP} is equal to $0.0441$ for $D_A$ over the  range $[0,6.7]$ and $2.1488$ for $D_H$ over the range $[0,11.5]$.
These values seem very good in the light of the fact that here the reconstruction is based on $18$ sample points against the $199$ points used in  Figures~\ref{fig:areaFFkriging} and~\ref{fig:hausFFkriging}. 

\begin{figure}[htbp]
%  \centering
%  \begin{minipage}[c]{0.48\textwidth}
%  \centering
%  \includegraphics[width=\textwidth]{lhsCCmaxHaus_2max.png}
%  \caption{Design points for $D_H$ used in the optimization in Figure~\ref{fig:lhsCCmaxHaus_poly0_2max}, when two local maxima points are selected in the second step with $\epsilon_{DIFF}=1$ $m$. The same symbols of Figure~\ref{fig:lhsCCmax_correct} have been used. }\label{fig:lhsCCmaxHaus_2max}
%  \end{minipage}%
%  \hspace{0.3cm}
%  \begin{minipage}[c]{0.48\textwidth}
%  \centering
%  \includegraphics[width=\textwidth]{lhsCCmaxHaus_poly0_2max.png}
%  \caption{Kriging optimization of $D_H$ based on two local maxima, with a constant regression term and gaussian correlation with $18$ points.}
%  \label{fig:lhsCCmaxHaus_poly0_2max}
%  \end{minipage}
 \centering
 \includegraphics[width=0.48\textwidth]{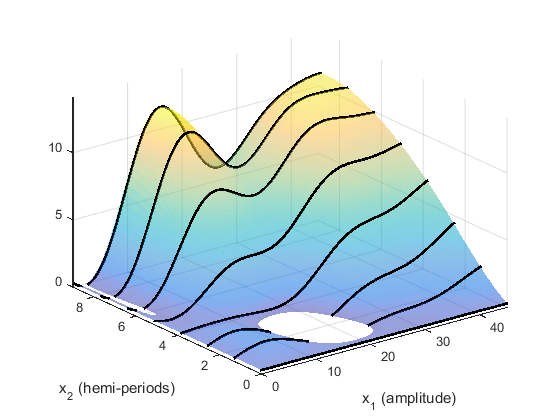}
 \caption{Kriging optimization of $D_H$ based on two local maxima, with a constant regression term and gaussian correlation with $18$ points.}
 \label{fig:lhsCCmaxHaus_poly0_2max}
 \end{figure}

However, if the selection of the design points for the second step is based on all evident local maxima in place of just one maximum, the optimization of $D_H$ depicted in Figure~\ref{fig:lhsCCmaxHaus_poly0_2max} improves notably. For example, if we fix $\epsilon_{DIFF}=1$ $m$, also the point $(40,9)$ is selected as a central point for the second step design whose points' coordinates are given in the fifth and sixth rows of Table~\ref{tab:designPoints}.
Three of the eight design points are near $(15,8)$, two near $(40,9)$ and the remaining three points are reallocated randomly. 
The accuracy of estimation of the maximum value in this case reduces to $ABS = 0.2540$. The same strategy applied to $D_A$ does not produce significant improvements.

%--------------------------------------------------------------

\subsection{Adaptive Design for Estimation}\label{subsec:simBP}
\begin{figure}[htbp]
\centering
\begin{minipage}[c]{0.48\textwidth}
\centering
\includegraphics[width=\textwidth]{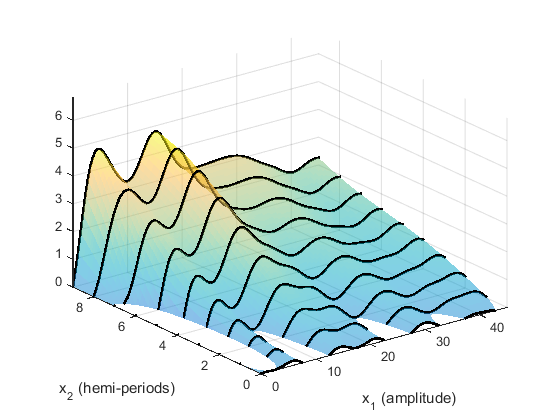}
\caption{Kriging reconstruction of $D_A$ with a quadratic regression term and gaussian correlation with $19$ points.}\label{fig:lhsCCmaxMSE_area_poly2}
\end{minipage}%
\hspace{0.3cm}
\begin{minipage}[c]{0.48\textwidth}
\centering
\includegraphics[width=\textwidth]{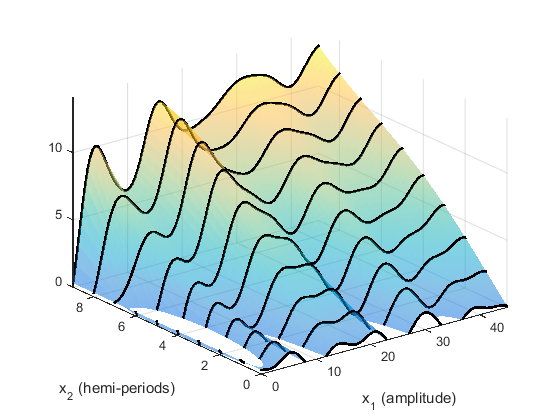}
\caption{Kriging reconstruction of $D_H$ with a quadratic regression term and gaussian correlation with $19$ points.}
\label{fig:lhsCCmaxMSE_haus_poly2}
\end{minipage}
\end{figure}

Following Subsection~\ref{subsec:BP}, the two-step procedure was tested for $D_A$ and $D_H$ and the same starting LHS of Section~\ref{subsec:simWP} was used.  
For both $D_A$ and $D_H$ the highest MSE value of the predictor was reached in $(2.5,9)$ which is very close to the border. 
Thus in the second step a point was placed in $(2.5,9)$, 
three points in its neighbourhood, and the remaining five points were allocated randomly. Their coordinates are in the last two rows of Table~\ref{tab:designPoints}. 
The results of the reconstructions on a regular $100 \times 100$ grid are in Figures~\ref{fig:lhsCCmaxMSE_area_poly2}  and~\ref{fig:lhsCCmaxMSE_haus_poly2} for $D_A$ and $D_H$, respectively. 
%  \begin{figure}[htbp]
%  \centering
%  \includegraphics[width=0.48\textwidth]{lhsCCmaxMSE_area.png}
%  \caption{Design points used in the reconstruction of $D_A$ and $D_H$ in Figures~\ref{fig:lhsCCmaxMSE_area_poly2} and~\ref{fig:lhsCCmaxMSE_haus_poly2}, respectively. The same symbols of Figure~\ref{fig:lhsCCmax_correct} have been used.
%  }
%  \label{fig:lhsCCmaxMSE}
%  \end{figure}
For $D_A$, the estimated covariance structure parameters are $\sigma^2= 1.5849$, $\theta_1 = 20$ and  $\theta_2 =0.5731$, while for $D_H$ they are $\sigma^2=  6.7214$, $\theta_1 = 20$ and  $\theta_2 =  0.5731$. 
Concerning the goodness of the reconstruction of $D_A$, the ABS errors 
%in Subsection~\ref{subsec:BP} 
are $\max ABS_{error}=2.0180$ and $\mean ABS_{error}=0.3536$, while for $D_H$ they are $\max ABS_{error}=3.8330$ and $\mean ABS_{error}=1.1212$.
Similar considerations to those in Section~\ref{subsec:simWP} apply here.

We conclude our example sections extending the discussion about the execution of target paths with small amplitude. 
As already observed when analyzing the border effect in Subsection~\ref{subsec:firstSim}, for very small amplitudes the performance indices are almost zero as expected for a vehicle of the dimensions of Charlie. 
Essentially the vehicle does not and cannot distinguish these paths from the straight line. This is evident in the $\mathcal R$ series that the controller transmits to the vehicle and is reflected in the reconstruction of $D_A$ and $D_H$. %, reflects 
When the amplitude of the target path is almost equal to the dimensions of the vehicle, there are low local maxima in the reconstructions witnessing the fact that the vehicle keeps  adjusting its position. 
We could conclude that the method in the paper is useful to evaluate these aspects of the  controller as well. 

%---------------------------------------------------------------------
\section{Conclusion}
\label{sec:concl}
This paper contributes to recent literature on establishing good experimental procedures and on defining performance indices when executing tasks along a curve 
and it presents a simulated study for path-following experiments for UMVs. Considerations throughout the paper are tailored to surface marine robots, but the methodology equally applies to mobile robots on land or underwater marine vehicles executing path-following tasks at constant depth.  
The paper addresses issues on how to define path-following experiments, how to evaluate the accuracy and efficiency of execution of a path and how to compare performances achieved by different vehicles or systems. 
Vehicle performance and capabilities are characterized in terms of absolute indices without reference to the physical characteristics of the vehicle.
The possibility of embedding the manoeuvrability restrictions and the difficulty of path execution is partially discussed in the paper, but still needs a thorough investigation.

The paper advocates the use of extensive simulations prior to the
execution of trials at sea and the use of statistically sound  methodologies from design of experiments for choosing experiments. Both are helpful for reducing the number of trials at sea, which are very costly, for increasing the quality of collected data, and for properly quantifying the loss of information with respect to larger experiments which might be unfeasible or useless for performances' comparison.   
Advanced feasibility studies for path-following are reported in the paper. They were instrumental in disproving wrong assumptions about the behaviour of UMVs and in getting a better understanding. The reported test were done with the protocol called {\em DeepRuler} and the Charlie USV simulator. 

Ultimately, it would be interesting to test the GEMs reported in the paper on other vehicles and data sets, simulated or not. Extensive test campaigns at sea are planned to properly interpret the performance indices and test the adaptive methodology under the effect of uncontrollable external conditions.  
A more sophisticated and automatic procedure for the selection of the design and for the modelling phase could be implemented in the {\em DeepRuler} software, e.g. following~\cite{challenor2013experimental}. Nevertheless these should remain as simple as possible also in the consideration of the expected extension to path-tracking and to online use.

\section*{Acknowledgments}
The authors are grateful to the other members of the CNR-ISSIA group for useful discussions, access to the simulator and sharing  their data sets. 

\bibliographystyle{model1-num-names}
\bibliography{IEEEbib.bib}

% %% Authors are advised to submit their bibtex database files. They are
% %% requested to list a bibtex style file in the manuscript if they do
% %% not want to use model1-num-names.bst.

% %% References without bibTeX database:

% % \begin{thebibliography}{00}

% %% \bibitem must have the following form:
% %%   \bibitem{key}...
% %%

% % \bibitem{}

% % \end{thebibliography}

\end{document}